\documentclass[10pt,twocolumn,letterpaper]{article}

\usepackage{cvpr}
\usepackage{times}
\usepackage{epsfig}
\usepackage{graphicx}
\usepackage{amsmath}
\usepackage{amssymb}

\usepackage{algorithm}
\usepackage{algorithmic}

\usepackage[pagebackref=true,breaklinks=true,letterpaper=true,colorlinks,bookmarks=false]{hyperref}

\cvprfinalcopy %

\def\cvprPaperID{0034} %

\ifcvprfinal\fi
\begin{document}

\title{Guided Anisotropic Diffusion and Iterative Learning for\\ Weakly Supervised Change Detection}

\author{Rodrigo Caye Daudt$^{1,2}$ \;\; Bertrand Le Saux$^{1}$ \;\; Alexandre Boulch$^{1}$ \;\;  Yann Gousseau$^{2}$\vspace{8pt}\\ 
		$^1$ ONERA  \;\;  
		$^2$ T\'{e}l\'{e}com ParisTech\vspace{5pt}\\ 
		{\tt\small [firstname.lastname]@[onera/telecom-paristech].fr}\vspace{-2pt}
	}

\maketitle

\begin{abstract}
Large scale datasets created from user labels or openly available data have become crucial to provide training data for large scale learning algorithms. While these datasets are easier to acquire, the data are frequently noisy and unreliable, which is motivating research on weakly supervised learning techniques. In this paper we propose an iterative learning method that extracts the useful information from a large scale change detection dataset generated from open vector data to train a fully convolutional network which surpasses the performance obtained by naive supervised learning. We also propose the guided anisotropic diffusion algorithm, which improves semantic segmentation results using the input images as guides to perform edge preserving filtering, and is used in conjunction with the iterative training method to improve results.
\end{abstract}

\section{Introduction}

Change detection (CD) is one of the oldest problems studied in the field of remote sensing image analysis~\cite{hussain2013change, singh1989review}. It consists of comparing a pair or sequence of coregistered images and identifying the regions where meaningful changes have taken place between the first and last acquisitions. However, the definition of meaningful change varies depending on the application. Changes of interest are, for example, new buildings and roads, forest fires, and growth or shrinkage of water bodies for environmental monitoring. Although exceptions exist, such as object based methods, most change detection algorithms predict a change label for each pixel in the provided images by modelling the task mathematically as a segmentation or clustering problem. 

Many variations of convolutional neural networks (CNNs)~\cite{lecun1998gradient}, notably fully convolutional networks (FCNs)~\cite{long2015fully}, have recently achieved excellent performances in change detection tasks~\cite{daudt2018fully, daudt2018hrscd, guo2018learning}. These methods require large amounts of training data to perform supervised training of the proposed networks~\cite{lecun2015deep}. Open labelled datasets for change detection are extremely scarce and are predominantly very small compared to labelled datasets in other computer vision areas. Benedek and Szirányi~\cite{benedek2009change} created the Air Change dataset which contain about 8 million labelled pixels, divided into three regions. Daudt \etal created the OSCD~\cite{daudt2018urban} dataset from Sentinel-2 multispectral images, with a total of about 9 million labelled pixels. While these datasets allow for simple models to be trained in a supervised manner, training more complex models with these data would lead to overfitting.

\begin{figure}[t]
    \begin{minipage}{0.32\linewidth}
        \centering
        \includegraphics[width=\linewidth]{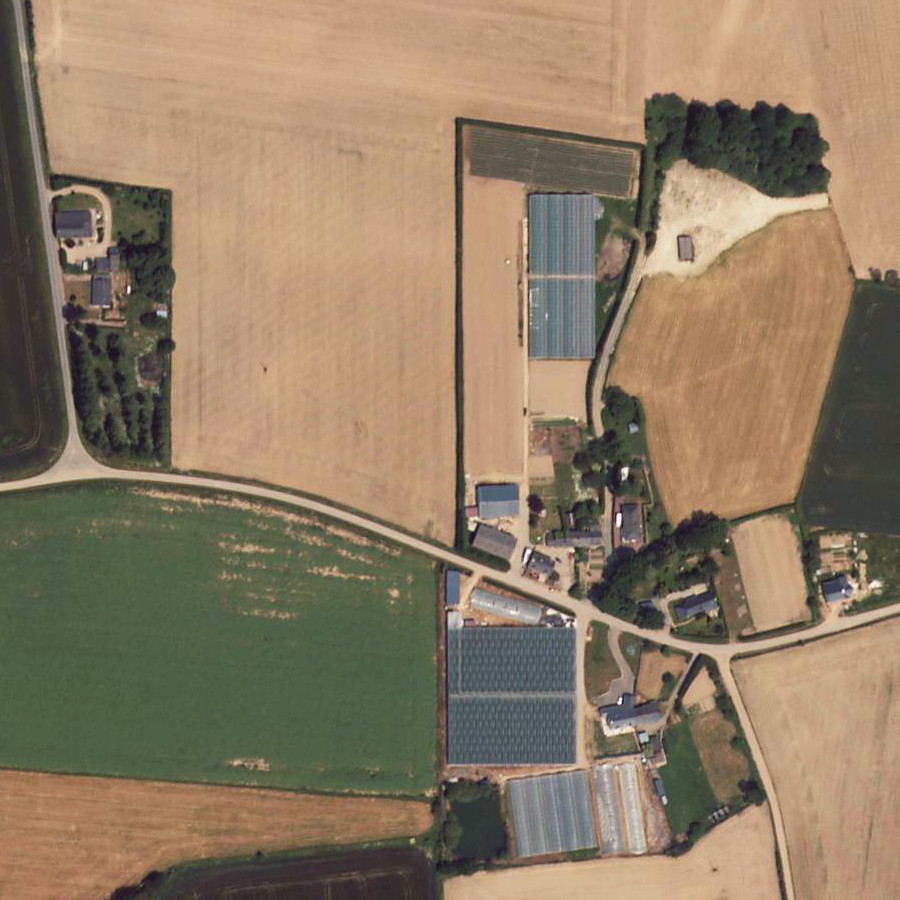}\\
        (a) Image 1
    \end{minipage}
    \hfill
    \begin{minipage}{0.32\linewidth}
        \centering
        \includegraphics[width=\linewidth]{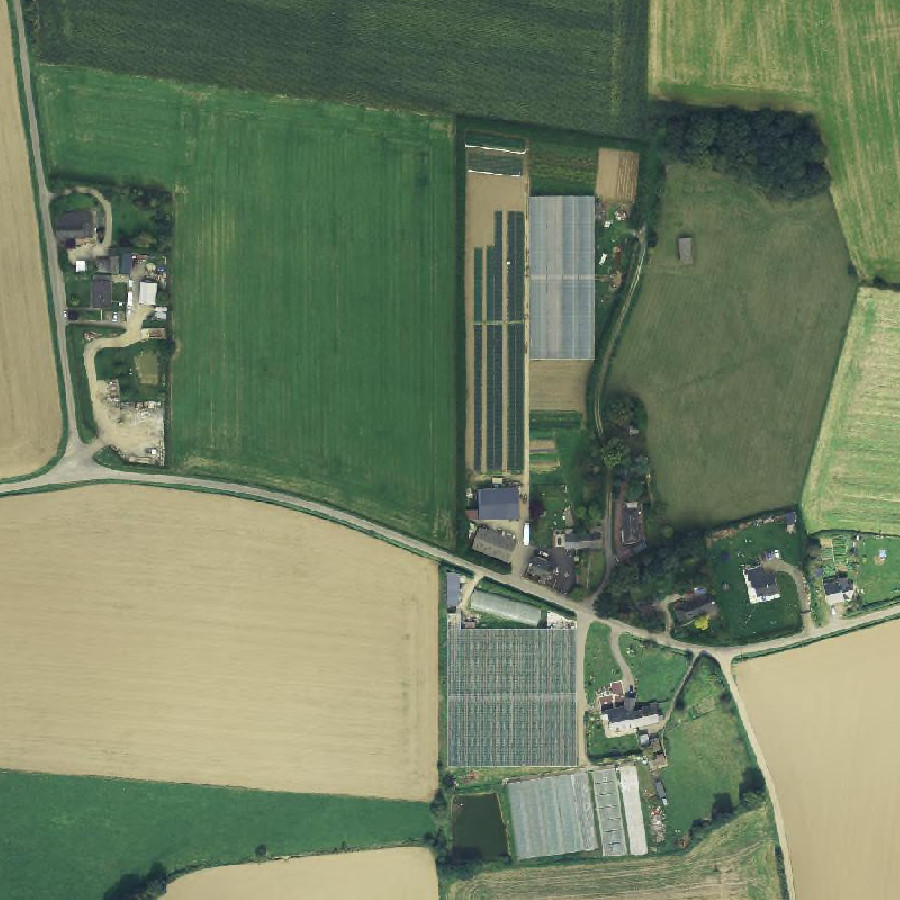}\\
        (b) Image 2
    \end{minipage}
    \hfill
    \begin{minipage}{0.32\linewidth}
        \centering
        \includegraphics[width=\linewidth]{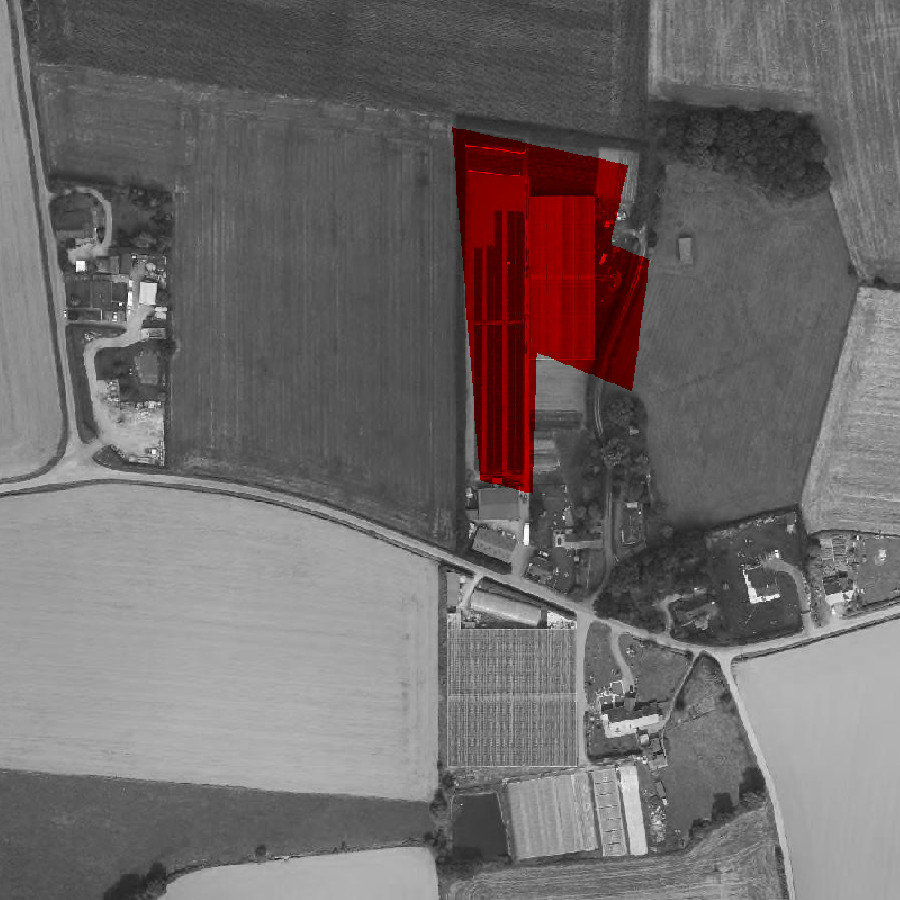}\\
        (c) Reference data
    \end{minipage}\break%
    
    \begin{minipage}{0.32\linewidth}
        \centering
        \includegraphics[width=\linewidth]{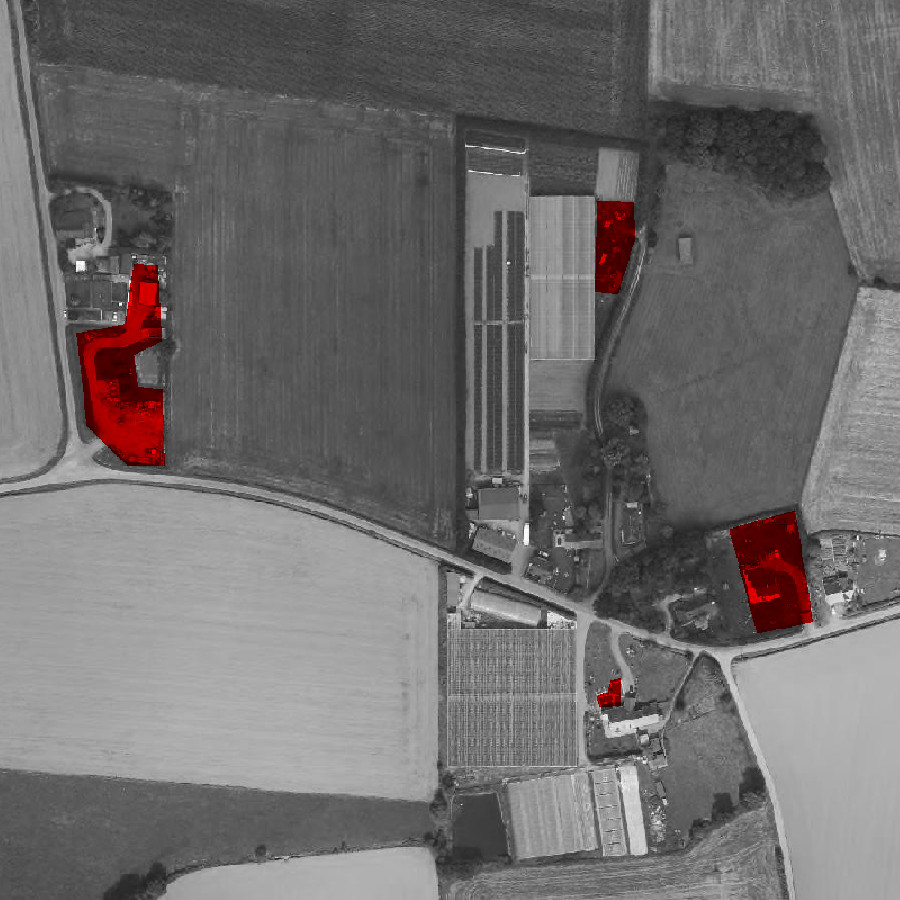}\\
        (d) Manual GT
    \end{minipage}
    \hfill
    \begin{minipage}{0.32\linewidth}
        \centering
        \includegraphics[width=\linewidth]{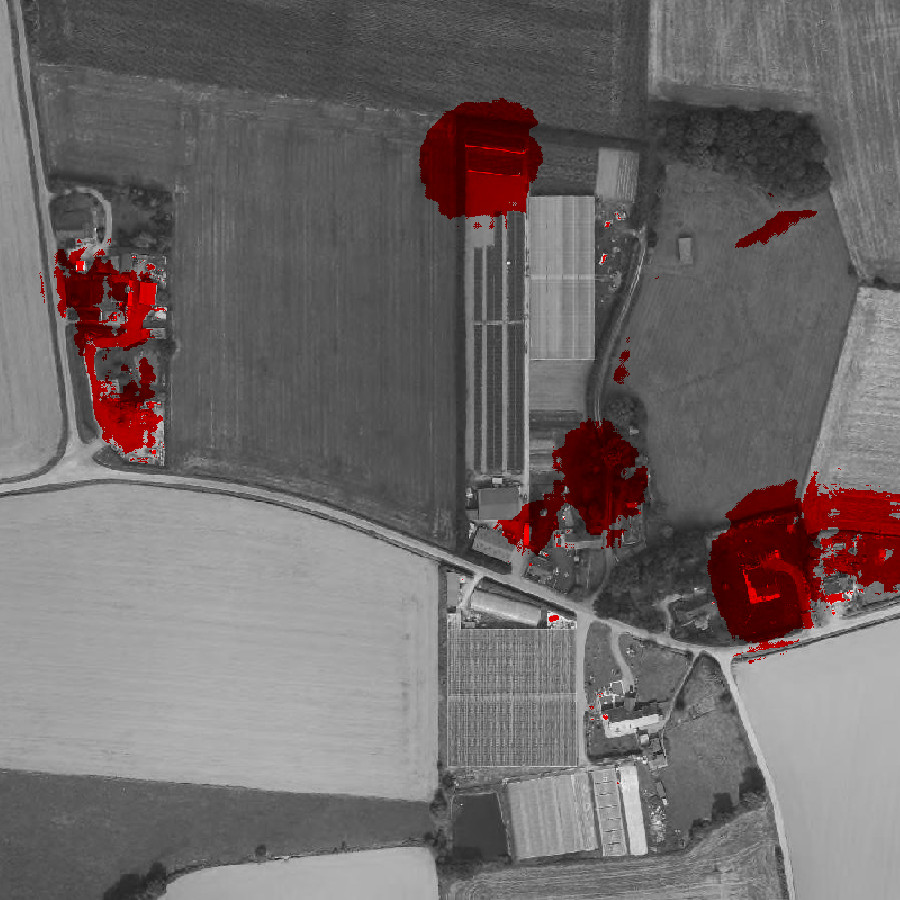}\\
        (e) Naive 
    \end{minipage}
    \hfill
    \begin{minipage}{0.32\linewidth}
        \centering
        \includegraphics[width=\linewidth]{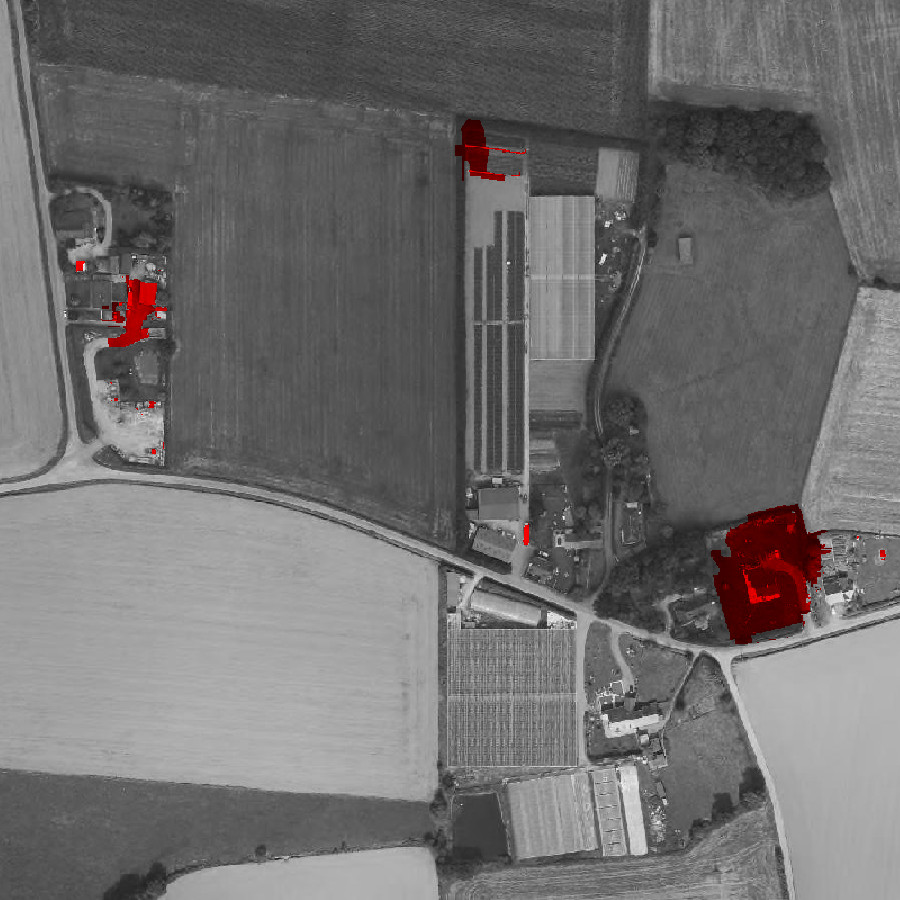}\\
        (f) Proposed
    \end{minipage}\break%
    
    \caption{(a)-(b) image pair, (c) change labels from the HRSCD dataset, (d) ground truth created by manually annotating changes, (e) result obtained by naive supervised training, (f) result obtained by our proposed method.}
    \label{fig:teaser}
\end{figure}

The recently proposed High Resolution Semantic Change Detection (HRSCD) dataset~\cite{daudt2018hrscd} is the first large scale change detection dataset. By combining an aerial image database with open change and land cover data, change maps and land cover maps were generated for almost 30 billion pixels, over 3000 times larger than previous change detection datasets. This dataset, however, contains unreliable labels due to having been generated automatically. The effect of naively using these data for supervised learning of change detection networks is shown in Fig.~\ref{fig:teaser}. Inaccuracies in the reference data stem primarily from two causes: imperfections in the vector data, and temporal misalignment between the annotations and the images. Naive supervision using such data leads to overestimation of the detected changes, as can be seen in Fig.~\ref{fig:teaser}(e). Nevertheless, there is much useful information in the available annotations that, if used adequately, can lead to better CD systems.

Due to the way the ground truth was generated, the labels in the dataset mark changes at a land parcel level with imprecise boundaries.
While useful for global monitoring of changes in land cover, it cannot delineate precise object-level changes.
In order to achieve a precise pixel-wise change detection, we propose a weakly supervised learning approach to change detection.
We consider the parcel-wise reference data as approximations, similar to bounding-boxes, of an ideal unknown ground truth corresponding to changes at pixel level. For each parcel with detected changes, the reference data in HRSCD contained both good and bad labels.
For this reason, the noise in the labels is not randomly distributed, but it is conditioned on the pixels' neighborhoods and highly structured.

We propose a weakly supervised approach to change detection that improves on previously proposed methods for semantic segmentation. We present a training scheme that harnesses the useful information in the HRSCD dataset for parcel-wise change detection, attempting to refine the reference data while training a fully convolutional network. By acknowledging the presence of incorrect labels in the training dataset (with respect to our fine grained objective), we are able to select good data and ignore bad ones, improving the final results as seen in Fig.~\ref{fig:teaser}(f). A preliminary version of this idea has been proposed in \cite{daudt2018learning}. This paper's new contributions include detailed equations and algorithms, integration with image-guided processing methods, and quantitative evaluation of the proposed methods.

This paper describes two main contributions to this problem. 
The first one is an iterative training scheme that alternates between training a fully convolutional network for change detection and using this network to find bad examples in the training set.
The second main contribution is the Guided Anisotropic Diffusion (GAD) algorithm, which is used in the iterative training scheme to better fit semantic segmentation predictions to the input images.
The proposed GAD algorithm is not restricted to change detection and can be used as a post-processing technique to improve semantic segmentation algorithms.

\section{Related Work}

\textbf{Change detection} has a long history, being one of the early problems tackled in remote sensing image understanding~\cite{singh1989review}. It is done using coregistered image pairs or sequences, and consists of identifying areas in the images that have experienced significant modifications between the acquisitions. Many of the state-of-the-art ideas in pattern recognition have been used for change detection in the past, from pixel-level comparison of images, to superpixel segmentation, object-level image analysis, and image descriptors~\cite{hussain2013change}. In this paper we treat change detection as a two class semantic segmentation problem, in which a label is predicted for each pixel in the input images. With the rise of machine learning algorithms for semantic segmentation, notably convolutional neural networks, many algorithms have attempted to learn to perform change detection. Most algorithms circumvented the problem of scarcity of training data through transfer learning by using pre-trained networks to generate pixel descriptors~\cite{sakurada2015change, el2016convolutional, el2017zoom}. Fully convolutional networks trained end-to-end to perform change detection have recently been proposed by several authors independently, usually using Siamese architectures~\cite{zhan2017change, daudt2018fully, daudt2018hrscd, chen2018mfcnet, guo2018learning}.

\textbf{Semantic segmentation} algorithms attempt to understand an input image and predict to which class among a known set of classes each pixel in an input image belongs. Change detection is modelled in this paper and many others as a semantic segmentation problem which takes as input two or more images. Long \etal proposed the first fully convolutional network for semantic segmentation, which achieved excellent performance and inference speed~\cite{long2015fully}. Since then, several improvements have been proposed for CNNs and FCNs. Ioffe and Szegedy have proposed batch normalization layers, which normalize activations and help avoid the vanishing/exploding gradient problem while training deep networks~\cite{ioffe2015batch}. Ronneberger \etal proposed the usage of skip connections that transfer details and boundary information from earlier to later layers in the network, which improves the accuracy around the edges between semantic regions~\cite{ronneberger2015u}. He \etal proposed the idea of residual connections, which have improved the performance of CNNs and FCNs and made it easier to train deep networks~\cite{he2016deep}.

\textbf{Noisy labels} for supervised learning is a topic that has already been widely explored~\cite{frenay2014comprehensive, frenay2014classification}. In many cases, label noise is completely random and independent from the data, and is modelled mathematically as such~\cite{natarajan2013learning, xiao2015learning, rolnick2017deep}. Rolnick \etal showed that supervised learning algorithms are robust to random label noise, and proposed strategies to further minimize the effect label noise has on training, such as increasing the training batch sizes~\cite{rolnick2017deep}. In the case presented in this paper, the assumption that the label noise is random does not hold. Incorrect change detection labels are usually around edges between regions or grouped together, which leads the network to learn to overestimate detected changes as seen in Fig.~\ref{fig:teaser}(e). Ignoring part of the training dataset, known as data cleansing (or cleaning), has already been proposed in different contexts~\cite{matic1992computer, john1995robust, guyon1996discovering, jeatrakul2010data}.

\textbf{Weakly supervised learning} is the name given to the group of machine learning algorithms that aim to perform different or more complex tasks than normally allowed by the training data at hand. Weakly supervised algorithms have recently gained popularity because they provide an alternative when data acquisition is too expensive. The problem of learning to perform semantic segmentation using only bounding box data or image level labels is closely related to the task discussed in this paper, since most methods propose the creation of an approximate semantic segmentation ground truth for training and dealing with its imperfections accordingly. Dai \etal proposed the BoxSup algorithm~\cite{dai2015boxsup} where region proposal algorithms are used to generate region candidates in each bounding box, then a semantic segmentation network is trained using these annotations, and finally it is used to select better region proposal candidates iteratively. Khoreva \etal proposed improvements to the BoxSup algorithm that includes using \textit{ad hoc} heuristics and an ignore class during training~\cite{khoreva2017simple}. They obtained best results using region proposal algorithms to create semantic segmentation training data directly from bounding boxes. Lu \etal modelled this problem as a simultaneous learning and denoising task through a  convex optimization problem~\cite{lu2017learning}. Ahn and Kwak proposed combining class activation maps, random walk and a learned network that predicts if pixels belong to the same region to perform semantic segmentation from image level labels~\cite{ahn2018learning}.

\textbf{Post-processing} methods that use information from guide images to filter other images, such as semantic segmentation results, have also been proposed~\cite{petschnigg2004digital, kopf2007joint, ferstl2013image}. A notable example is the Dense CRF algorithm proposed by Kr\"ahenb\"uhl and Koltun, in which an efficient solver is proposed for fully connected conditional random fields with Gaussian edge potentials~\cite{krahenbuhl2011efficient}. The idea of using a guide image for processing another is also the base of the Guided Image Filtering algorithm proposed by He \etal~\cite{he2013guided}, where a linear model that transforms a guide image into the best approximation of the filtered image is calculated, thus transferring details from the guide image to the filtered image. The use of joint filtering is popular in the field of computational photography, and has been used for several applications~\cite{petschnigg2004digital, kopf2007joint, ferstl2013image}.
One of the building blocks of the filtering method we propose in this paper is the anisotropic diffusion, proposed by Perona and Malik~\cite{perona1990scale}, an edge preserving filtering algorithm in which the filtering of an image is modelled as a heat equation with a different diffusion coefficient at each edge between neighbouring pixels depending on the local geometry and contrast.
However, to the best of our knowledge, this algorithm has not yet been used for guided filtering.

\section{Method}\label{sec:method}

\begin{figure}[t]
    \centering
    \includegraphics[width=\linewidth]{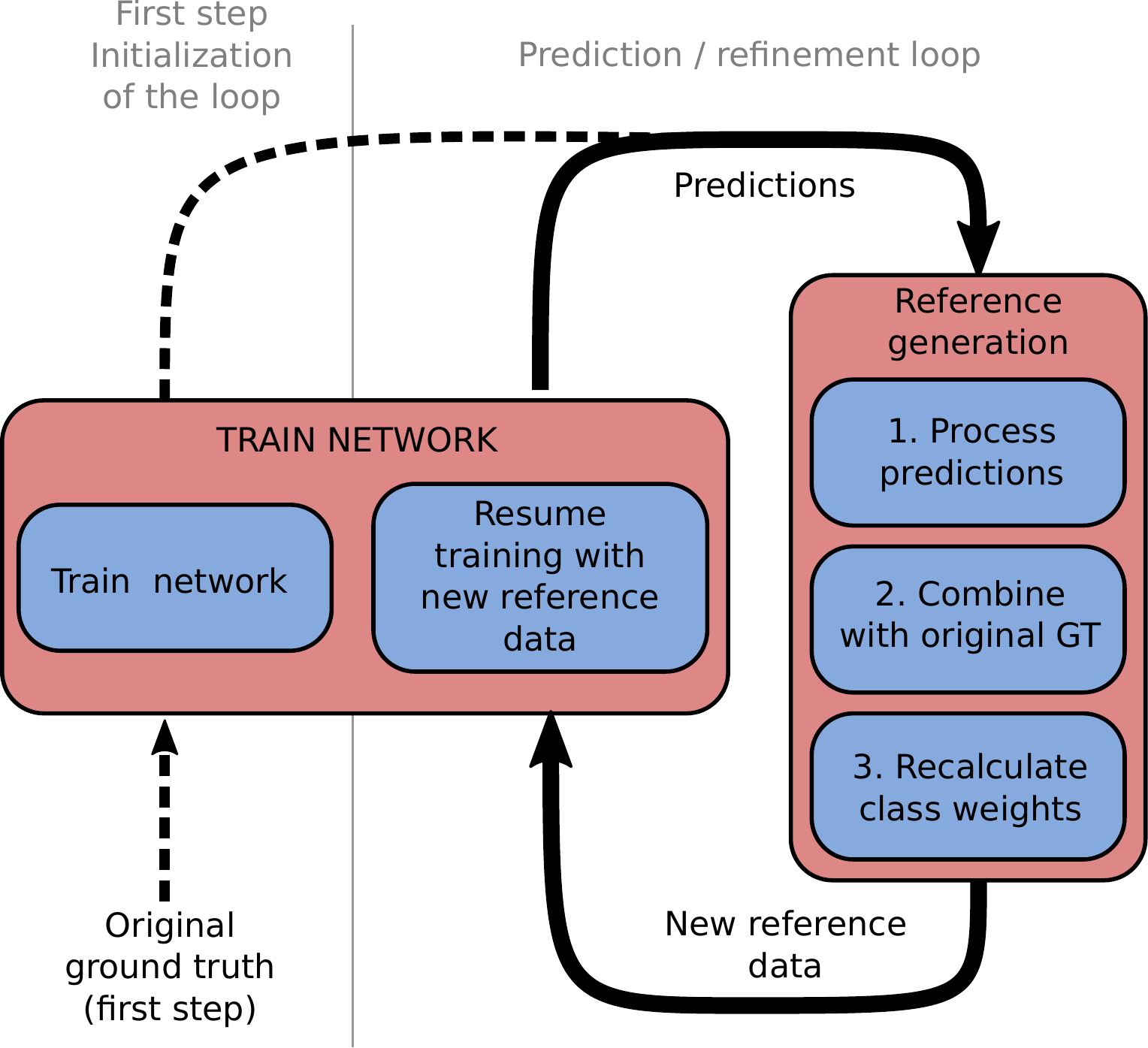}\\
    \caption{Iterative training method: alternating between training and data cleaning allows the network to simultaneously learn the desired task and to remove bad examples from the training dataset.}
    \label{fig:it_schematic}
\end{figure}

The two main contributions of this paper are: 1) an iterative training scheme that aims to efficiently learn from inaccurate and unreliable ground truth semantic segmentation data and 2) the guided anisotropic diffusion algorithm, which uses information from the input images to filter and improve semantic segmentation results.  These contributions are presented in sections \ref{sec:its} and \ref{sec:gad} below, respectively. While these two ideas are presented in this paper in the context of change detection, the proposed methods' scope is broader and could be used for other semantic segmentation problems, together or separately.

\subsection{Iterative Training Scheme}
\label{sec:its}

The label noise present in the HRSCD dataset for change detection is challenging due to its spatial structure and correlation between neighbors. In the taxonomy presented in \cite{frenay2014comprehensive, frenay2014classification}, this type of label noise would be classified as "label noise not at random" (NNAR). NNAR is the most complex among the label noise models in the taxonomy.
In the case of HRSCD, most errors can be attributed to one of the following reasons: the available information is insufficient to perform labelling, errors on the part of the annotators, subjectiveness of the labelling task, and temporal misalignment between the databases used to create the HRSCD dataset.

It is important to note that, as discussed by Fr{\'e}nay and Kab{\'a}n in \cite{frenay2014comprehensive}, label noise has an even more powerful damaging impact when a dataset is imbalanced since it alters the perceived, but not the real, class imbalance and therefore the methods used to mitigate class imbalance during training are less effective. In the case of change detection with the HRSCD dataset, the no change class outnumbers the change class 130 to 1, which means the label noise could significantly alter the calculated class weights used for training.

It has been noticed in \cite{daudt2018hrscd} and in our own experiments that change detection networks trained directly on the HRSCD dataset had the capacity to detect changes in image pairs but tended to predict blobs around the detected change instances, as is depicted in Fig.~\ref{fig:gad_122}(c), likely in an attempt to minimize the loss for the training images where the surrounding pixels of true changes are also marked as having experienced changes. In many cases, it was observed that the network predictions were correct where the ground truth labels were not. Based on this observation, we propose a method for training the network that alternates between actual minimization of a loss function and using the network predictions to clean the reference data before continuing the training. A schematic that illustrates the main ideas of this method is shown in Fig.~\ref{fig:it_schematic}. For the remainder of this paper, the iteration cycles of training the network and cleaning of training data will be referred to as \textit{hyperepochs}.

\begin{algorithm}[t]
    \caption{Iterative training pseudocode.}
    \begin{algorithmic}[1]
        \STATE \textbf{Input:} $I$: Image pairs, $GT_o$: Original unreliable ground truths, $N$: Number of hyperepochs, $\Phi_r$: Initial random network weights.
        \STATE \textbf{Output:} $\Phi_N$: Trained network weights.
        \STATE $w_0 \gets$ calculate class weights inversely proportional to number of class examples
        \STATE $\Phi_0 \gets$ Train network with $I$ and $GT_0$ until convergence or fixed number of epochs
        \FOR{($i\gets1$; $i \leq N$; $i++$)}
            \STATE $P_i \gets$ generate predictions for training dataset with current network
            \STATE $P_{i,pp} \gets$ Post-processing of predictions\;
            \STATE $GT_i \gets$ Combine $P_{i,pp}$ with $GT_0$ to generate cleaner ground truth data
            \STATE $\Phi_i \gets$ Continue training network from $\Phi_{i-1}$ using $I$ and $GT_i$ until convergence
        \ENDFOR
    \end{algorithmic}
\end{algorithm}

Alternating between training a semantic segmentation network and using it to make changes to the training data has already been explored~\cite{dai2015boxsup, khoreva2017simple}. Such iterative methods are named "classification filtering"~\cite{frenay2014classification}. The main differences between the method proposed in this paper and previous ones are:
\begin{enumerate}
    \item \textbf{No bounding box information is available}: we work directly with pixel level annotations, which were generated form vector data;
    \item \textbf{Each annotated region may contain more than one instance}: the annotations often group several change instances together;
    \item \textbf{Annotations are not flawless}: the HRSCD dataset contains both false positives and false negatives in change annotations.
\end{enumerate}

It has also been shown by Khoreva \etal in \cite{khoreva2017simple} that simply using the outputs of the network as training data leads to degradation of the results, and that it is necessary to use priors and heuristics specific to the problem at hand to prevent a degradation in performance. In this paper we use two ways to avoid degradation of the results with iterative training. The first is using processing techniques that bring information from the input images into the predicted semantic segmentations, improving the results and providing a stronger correlation between inputs and predictions. The Guided Anisotropic Diffusion algorithm presented in Section~\ref{sec:gad} serves this purpose, but other algorithms such as Dense CRF~\cite{krahenbuhl2011efficient} may also be used. The second way the degradation of results is avoided is by combining network predictions with the original reference data at each iteration, instead of simply using predictions as reference data.

\begin{figure}
    \centering
    \begin{minipage}{0.3\linewidth}
        \centering
        \hspace{20pt}Orig. GT
        
        \rotatebox[origin=c]{90}{Pred. \hspace{8pt}}
        \begin{tabular}{c|c|c}
              & \textbf{0} & \textbf{1} \\ \hline
            \textbf{0} & 0 & 0 \\ \hline
            \textbf{1} & 0 & 1 \\
        \end{tabular}
        
        (a)~Intersection
    \end{minipage}
    \begin{minipage}{0.3\linewidth}
        \centering
        \hspace{20pt}Orig. GT
        
        \rotatebox[origin=c]{90}{Pred. \hspace{8pt}}
        \begin{tabular}{c|c|c}
              & \textbf{0} & \textbf{1} \\ \hline
            \textbf{0} & 0 & 2 \\ \hline
            \textbf{1} & 0 & 1 \\
        \end{tabular}
        
        (b)~FN$\gets$ Ignore
    \end{minipage}
    \begin{minipage}{0.36\linewidth}
        \centering
        \hspace{20pt}Orig. GT
        
        \rotatebox[origin=c]{90}{Pred. \hspace{8pt}}
        \begin{tabular}{c|c|c}
              & \textbf{0} & \textbf{1} \\ \hline
            \textbf{0} & 0 & 2 \\ \hline
            \textbf{1} & 2 & 1 \\
        \end{tabular}
        
        (c)~FN$\cup$FP$\gets$~Ignore
    \end{minipage}
    \label{ignore_strats}
    \vspace{6pt}
    \caption{Proposed methods for merging original labels and network predictions. Classes: 0 is no change, 1 is change, 2 is ignore. (a) Intersection between original and detected changes. (b) Ignore false negatives from the perspective of original labels. (c) Ignore all pixels with label disagreements.}
\end{figure}

\begin{figure}[t]
    \begin{minipage}{0.32\linewidth}
        \centering
        \includegraphics[width=\linewidth]{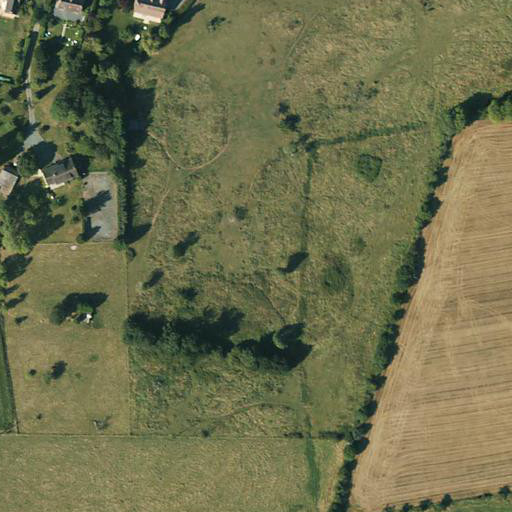}\\
        (a) Image 1
    \end{minipage}
    \hfill
    \begin{minipage}{0.32\linewidth}
        \centering
        \includegraphics[width=\linewidth]{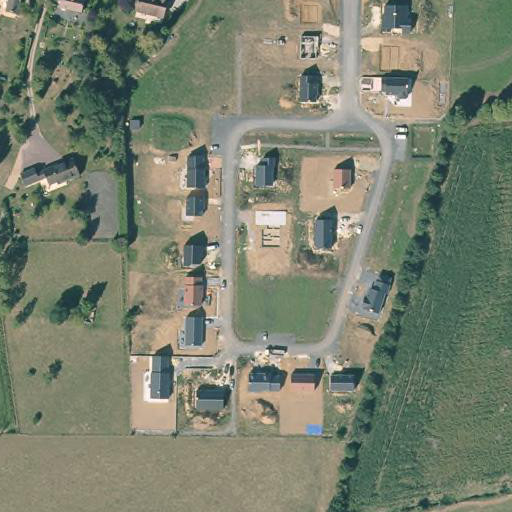}\\
        (b) Image 2
    \end{minipage}
    \hfill
    \begin{minipage}{0.32\linewidth}
        \centering
        \includegraphics[width=\linewidth]{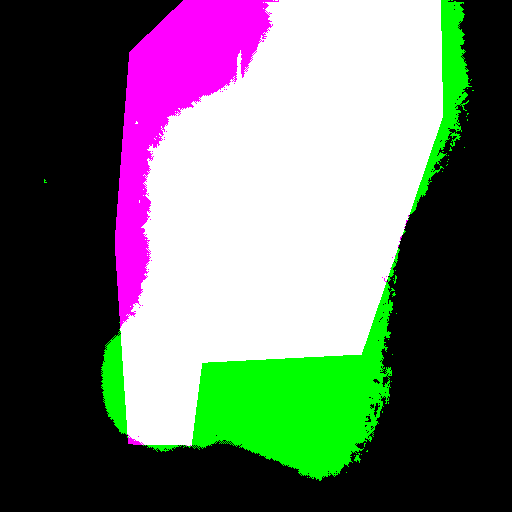}\\
        (c) GT and pred.
    \end{minipage}\break%
    
    \begin{minipage}{0.32\linewidth}
        \centering
        \includegraphics[width=\linewidth]{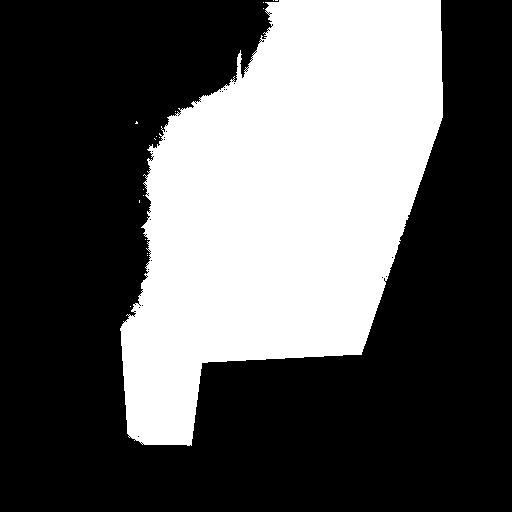}\\
        (d)~Intersection
    \end{minipage}
    \hfill
    \begin{minipage}{0.32\linewidth}
        \centering
        \includegraphics[width=\linewidth]{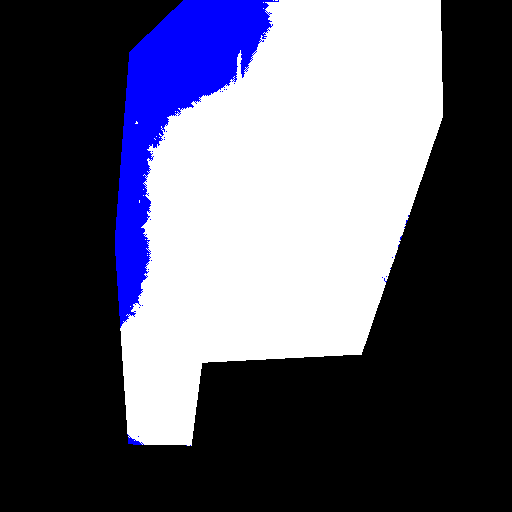}\\
        (e)~FN$\gets$ Ignore
    \end{minipage}
    \hfill
    \begin{minipage}{0.32\linewidth}
        \centering
        \includegraphics[width=\linewidth]{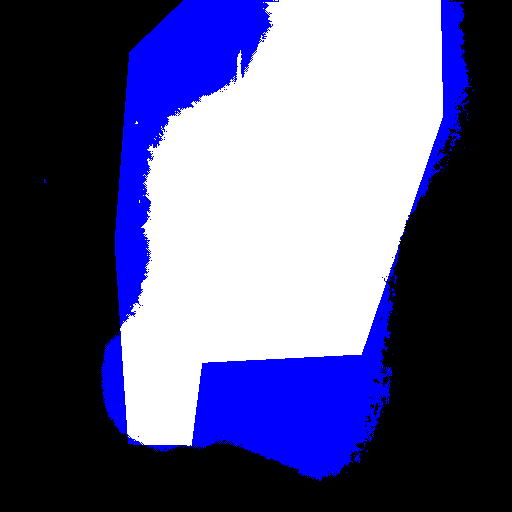}\\
        (f)~FN$\cup$FP$\gets$~Ign.
    \end{minipage}\break%
    
    \caption{Example case of the three proposed merge strategies. In (c), black is true negative, white is true positive, magenta is false negative, and green is false positive. In (d)-(f) blue represents the ignore class.}
    \label{fig:merge_example}
\end{figure}

\begin{figure}[t]
    \begin{minipage}{0.32\linewidth}
        \centering
        \includegraphics[width=\linewidth]{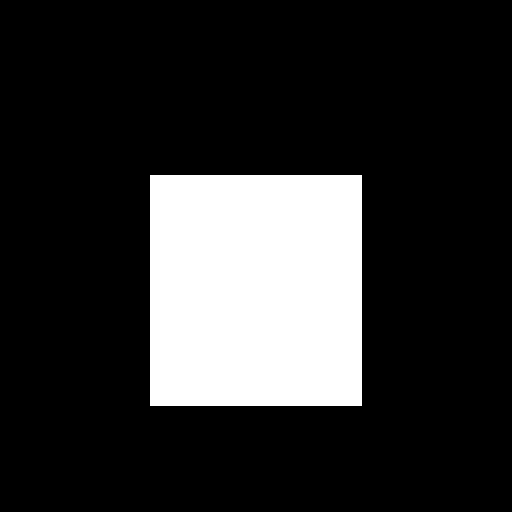}\\
        (a) Guide image
    \end{minipage}
    \hfill
    \begin{minipage}{0.32\linewidth}
        \centering
        \includegraphics[width=\linewidth]{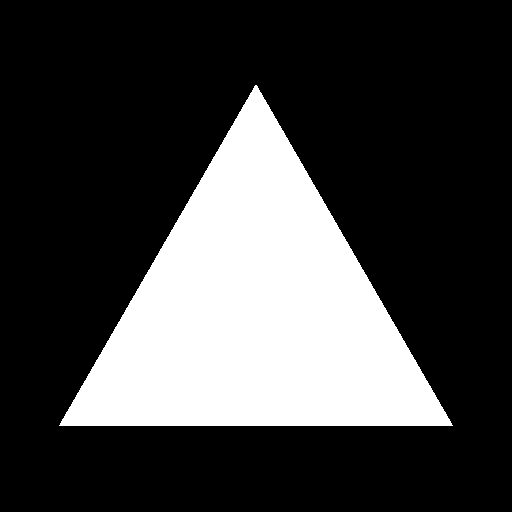}\\
        (b) Input image
    \end{minipage}
    \hfill
    \begin{minipage}{0.32\linewidth}
        \centering
        \includegraphics[width=\linewidth]{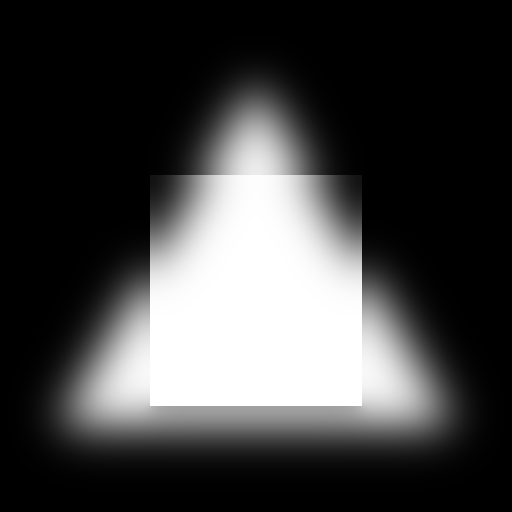}\\
        (c) 1000 it.
    \end{minipage}\break%
    
    \begin{minipage}{0.32\linewidth}
        \centering
        \includegraphics[width=\linewidth]{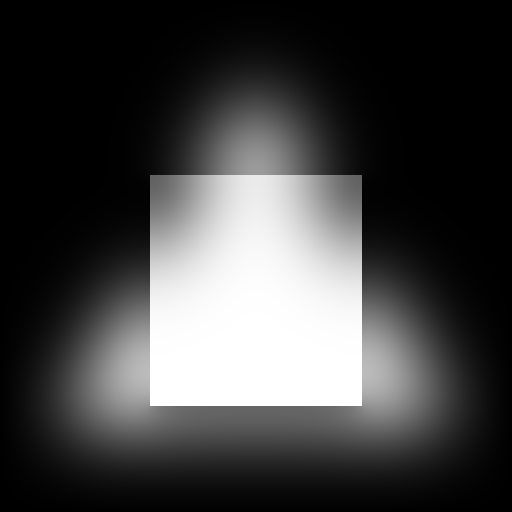}\\
        (d) 3000 it.
    \end{minipage}
    \hfill
    \begin{minipage}{0.32\linewidth}
        \centering
        \includegraphics[width=\linewidth]{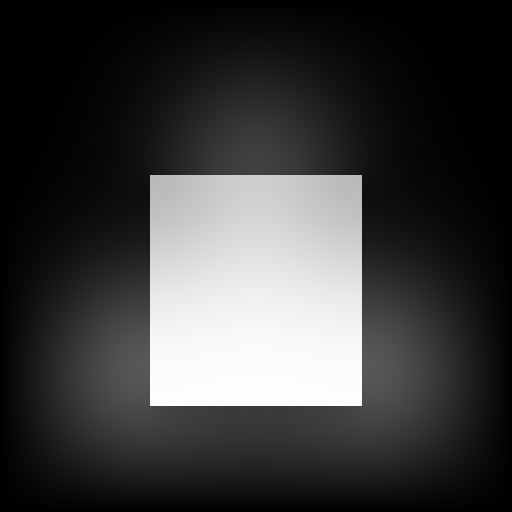}\\
        (e) 10000 it.
    \end{minipage}
    \hfill
    \begin{minipage}{0.32\linewidth}
        \centering
        \includegraphics[width=\linewidth]{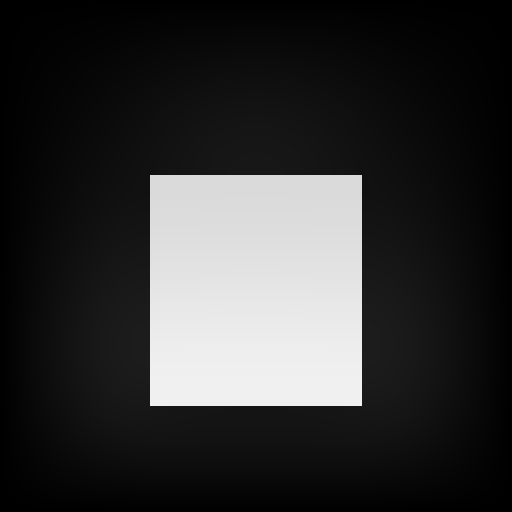}\\
        (f) 30000 it.
    \end{minipage}\break%
    
    \caption{Results of guided anisotropic diffusion. Edges in the guide image (a) are preserved in the filtered image (b). (c)-(f) show results using different numbers of iterations.}
    \label{fig:gad_triangle}
\end{figure}

We propose three ways of merging the original labels with network predictions. When merging, each pixel will have a binary label from the original ground truth and a binary label from the network prediction. If these labels agree, there is no reason to believe the label for that pixel is wrong, and it is therefore kept unchanged. In case the labels disagree, the following options to decide the pixel's label are proposed:
\begin{enumerate}
    \item \textbf{The intersection of predicted and reference change labels is kept as change}: this strategy assumes all changes are marked in both the reference data and in the prediction. It also puts pixels with uncertain labels in the no change class, where they are more easily diluted during training due to the class imbalance.
    \item \textbf{Ignore false negatives}: using an ignore class for false negatives attempts to keep only good examples in the change class, improving the quality of the training data. It assumes all changes are marked in the original labels provided.
    \item \textbf{Ignore all disagreements}: marking all label disagreements to be ignored during training attempts to keep only clean labels for training at the cost of reducing the number of training examples. This approach is the only one that is class agnostic.
\end{enumerate}
In practice, the ignored pixels are marked as a different class that is given a class weight of 0 during the training. Tables for the three proposed methods can be found in Fig.~\ref{ignore_strats}, and an example can be found in Fig.~\ref{fig:merge_example}.

\subsection{Guided Anisotropic Diffusion}
\label{sec:gad}

In their seminal paper, Perona and Malik proposed an anisotropic diffusion algorithm with the aim of performing scale space image analysis and edge preserving filtering~\cite{perona1990scale}. Their diffusion scheme has the ability to blur the inside of regions with homogeneous colours while preserving or even enhancing edges. This is done by modelling the filtering as a diffusion equation with spatially variable coefficients, and as such is an extension of the linear heat equation, whose solution is mathematically equivalent to Gaussian filtering when diffusion coefficients are constant~\cite{koenderink1984structure}. Diffusion coefficients are set to be higher where the local contrast of the image is lower.  

More precisely, we consider the anisotropic diffusion equation
\begin{equation}
    \frac{\partial I}{\partial t} = div(c(x,y,t)\nabla I) = c(x,y,t)\Delta I + \nabla c \cdot \nabla I
\end{equation}
where $I$ is the input image, $c(x,y,t)$ is the coefficient diffusion at position $(x,y)$ and time $t$, $div$ represents the divergence, $\nabla$ represents the gradient, and $\Delta$ represents the Laplacian. In its original formulation, $c(x,y,t)$ is a function of the input image I. To perform edge preserving filtering, one approach is using the coefficient
\begin{equation}
    c(x,y,t) = \frac{1}{1 + \Big(\frac{||\nabla I(x,y,t)||}{K}\Big)^{2}},
\end{equation}
which approaches $1$ (strong diffusion) where the gradient is small, and approaches 0 (weak diffusion) for large gradient values. Other functions with these properties and bound in $[0,1]$ may also be used. The parameter $K$ controls the sensitivity to contrast in the image.

In the guided anisotropic diffusion algorithm the aim is to perform edge preserving filtering on an input image, but instead of preserving the edges in the filtered image we preserve edges coming from a separate guide image (or images). Doing so allows us to transfer properties from the guide image $I_g$ into the filtered image $I_f$. An illustrative example is shown in Fig.~\ref{fig:gad_triangle}, where the image of a rectangle (a) is used as a guide to filter the image of a triangle (b). The edges from the guide image $I_g$ are used to calculate $c(x,y,t)$, which in practice creates barriers in the diffusion of the filtered image $I_f$, effectively transferring details from $I_g$ to $I_f$. These edges effectively separate the image in two regions, inside and outside the rectangle, and the gray values in each of these regions experience diffusion, but there is virtually no diffusion happening between them.

\begin{algorithm}[t]
    \caption{Guided Anisotropic Diffusion pseudocode.}
    \label{alg:gad}
    \begin{algorithmic}[1]
        \STATE \textbf{Input:}{$I_1$, $I_2$, $I_in$, $N$, $K$, $\lambda$}
        \STATE \textbf{Output:}{$I_f$}
        \STATE $I_f \gets I_in$
        \FOR {($i\gets1$; $i \leq N$; $i++$)}
            \FOR {($I_j = \{I_1, I_2\}$)}
                \STATE $\nabla I_j \gets$ Calculate gradient of $I_j$\;
                \STATE $c_{I_j} \gets$ Calculate using Eq.~\ref{eq:c}\;
                \STATE $I_j \gets I_j + \lambda \cdot \nabla I_j \cdot c_{I_j}$\;
            \ENDFOR
            \STATE $\nabla I_f \gets$ Calculate gradient of $I_f$\;
            \STATE $c_f \gets$ Calculate using Eq.~\ref{eq:multi_c}\;
            \STATE $I_f \gets I_f + \lambda \cdot \nabla I_f \cdot c_f$\;
        \ENDFOR
    \end{algorithmic}
\end{algorithm}

Our aim is to use this guided anisotropic diffusion (GAD) algorithm to improve semantic segmentation results based on the input images. Given that the change detection networks trained on the HRSCD dataset have the tendency to overestimate the area of the detected changes, GAD provides a way to improve these semantic segmentation results by making them more precisely fit the edges present in the input images. A few design choices were made to extend the anisotropic diffusion from gray level images to RGB image pairs. The extension to RGB image was done by taking the mean of the gradient norm at each location
\begin{equation}\label{eq:c}
    c_I(x,y,t) = \frac{1}{1 + \Big(\sum_{C \in \{R,G,B\}}\frac{||\nabla I_C(x,y,t)||}{3\cdot K}\Big)^{2}},
\end{equation}
so that edges in any of the color channels would prevent diffusion in the filtered image. To extend this further to be capable of taking multiple guide images simultaneously, which is necessary for the problem of change detection, the minimum diffusion coefficient at each position $(x,y,t)$ was used, once again to ensure that any edge present in any guide image would be transferred to the filtered image:
\begin{equation}\label{eq:multi_c}
    c_{I_1,I_2}(x,y,t) = min_{i \in \{1,2\}} c(I_i)(x,y,t).
\end{equation}

Guided anisotropic diffusion aims to improve semantic segmentation predictions by filtering the class probabilities yielded by a fully convolutional network. It is less adequate to correct for large classification mistakes, as opposed to non-local methods such as Dense CRF, but it leads to smoother predictions with more accurate edges. It can also be easily extended for any number of guide images by increasing the number of images considered in Eq.~\ref{eq:multi_c}. The pseudocode for the GAD algorithm can be found in Alg.~\ref{alg:gad}. As mentioned in the original anisotropic diffusion paper, the algorithm is unstable for $\lambda > 0.25$ when using 4-neighborhoods for the calculations.

\section{Experiments}\label{sec:experiments}

To validate the methods proposed in Section~\ref{sec:method} we adopted the hybrid change detection and land cover mapping fully convolutional network presented in \cite{daudt2018hrscd}, since it was already proven to work with the HRSCD dataset. We adopted \textit{strategy 4.2} described in the paper, in which the land cover mapping branches of the network are trained before the change detection one to avoid setting a balancing hyperparameter. The land cover mapping branches of the network were fixed to have the same parameter weights for all tests presented in this paper, and evaluating those results is not done here as the scope of this paper is restricted to the problem of change detection.

\begin{figure}[t]
    \begin{minipage}{0.32\linewidth}
        \centering
        \includegraphics[width=\linewidth]{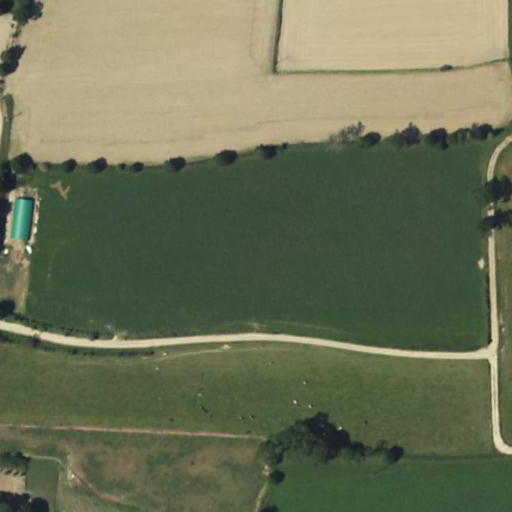}\\
        (a) Image 1
    \end{minipage}
    \hfill
    \begin{minipage}{0.32\linewidth}
        \centering
        \includegraphics[width=\linewidth]{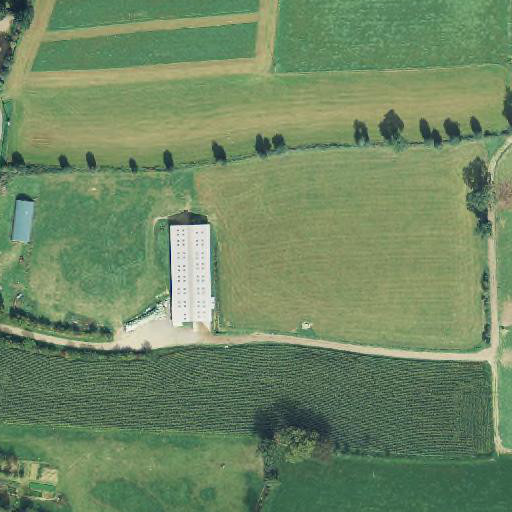}\\
        (b) Image 2
    \end{minipage}
    \hfill
    \begin{minipage}{0.32\linewidth}
        \centering
        \includegraphics[width=\linewidth]{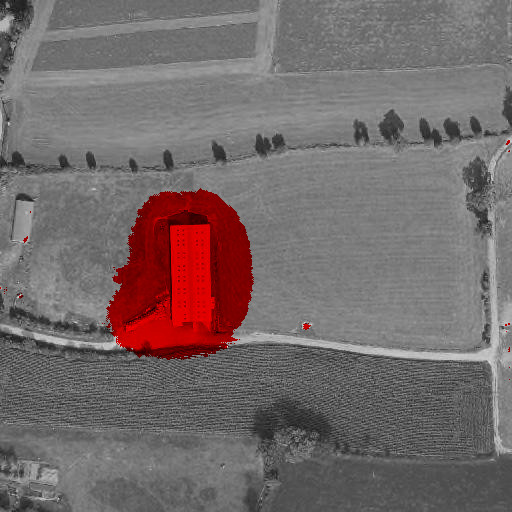}\\
        (c) Naive pred.
    \end{minipage}\break%
    
    \begin{minipage}{0.32\linewidth}
        \centering
        \includegraphics[width=\linewidth]{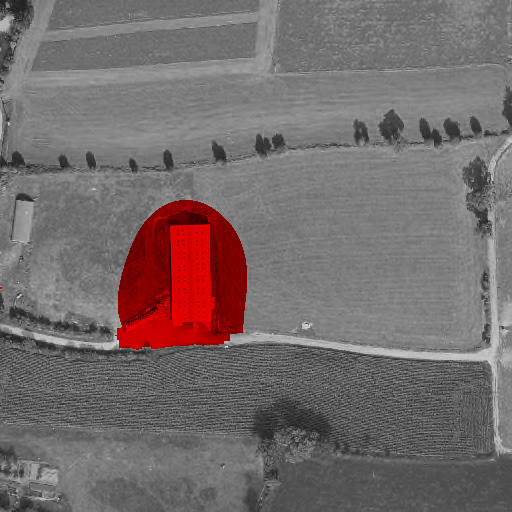}\\
        (d) 2000 it.
    \end{minipage}
    \hfill
    \begin{minipage}{0.32\linewidth}
        \centering
        \includegraphics[width=\linewidth]{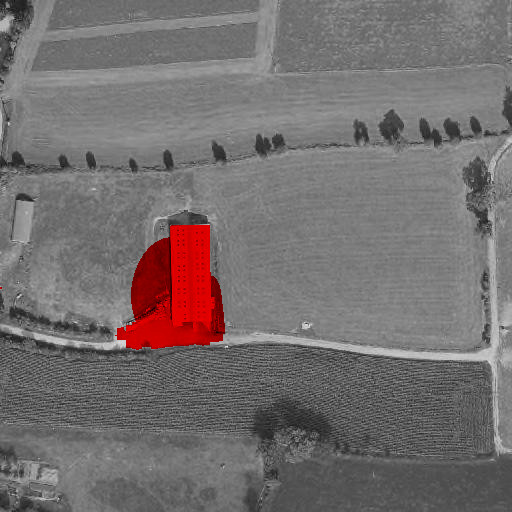}\\
        (e) 5000 it.
    \end{minipage}
    \hfill
    \begin{minipage}{0.32\linewidth}
        \centering
        \includegraphics[width=\linewidth]{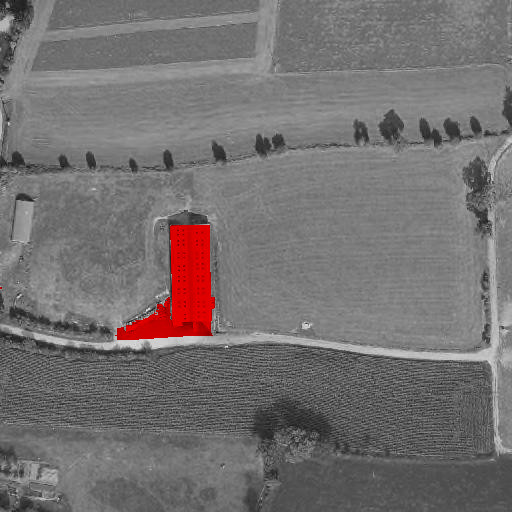}\\
        (f) 20000 it.
    \end{minipage}\break%
    
    \caption{Guided anisotropic diffusion for filtering a real example of semantic segmentation. The diffusion allows edges from the guide images to be transferred to the target image, improving the results.}
    \label{fig:gad_122}
\end{figure}

\begin{figure}[t]
    \begin{minipage}{0.32\linewidth}
        \centering
        \includegraphics[width=\linewidth]{ad_ex_122-01-I1.jpg}\\
        (a) Image 1
    \end{minipage}
    \hfill
    \begin{minipage}{0.32\linewidth}
        \centering
        \includegraphics[width=\linewidth]{ad_ex_122-02-I2.jpg}\\
        (b) Image 2
    \end{minipage}
    \hfill
    \begin{minipage}{0.32\linewidth}
        \centering
        \includegraphics[width=\linewidth]{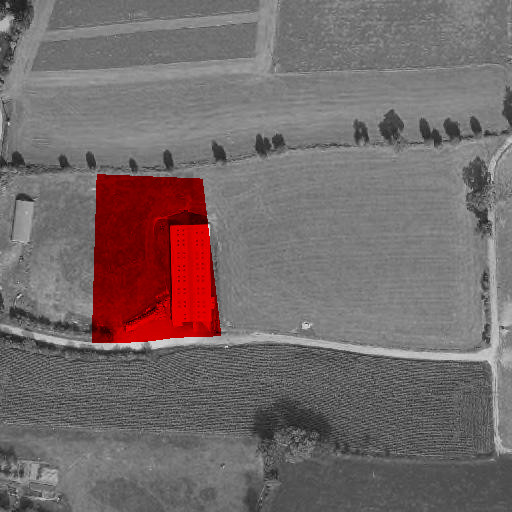}\\
        (c) Reference data
    \end{minipage}
    \break
    
    \begin{minipage}{0.32\linewidth}
        \centering
        \includegraphics[width=\linewidth]{ad_ex_122-03-pred-00000.jpg}\\
        (d) Naive pred.
    \end{minipage}
    \hfill
    \begin{minipage}{0.32\linewidth}
        \centering
        \includegraphics[width=\linewidth]{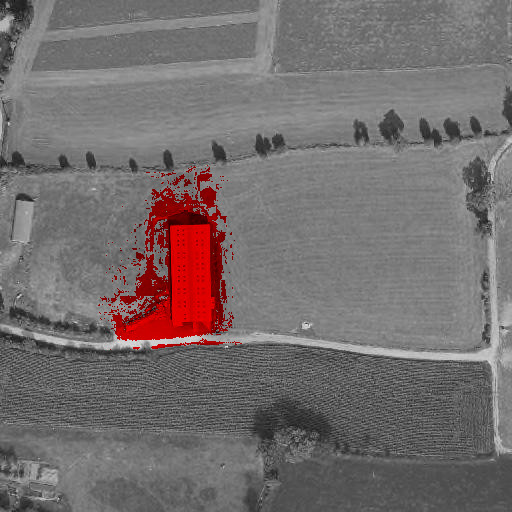}\\
        (e) Dense CRF
    \end{minipage}
    \hfill
    \begin{minipage}{0.32\linewidth}
        \centering
        \includegraphics[width=\linewidth]{ad_ex_122-06-pred-20000.jpg}\\
        (f) GAD
    \end{minipage}
    \break%
    
    \caption{Comparison between (c) original dataset ground truth, (e) prediction filtered by Dense CRF, and (f) prediction filtered with guided anisotropic diffusion for 20000 iterations.}
    \label{fig:gad_vs_dcrf}
\end{figure}

\begin{figure}[t]
    \centering
    \begin{minipage}{0.8\linewidth}
        \centering
        \includegraphics[width=\linewidth]{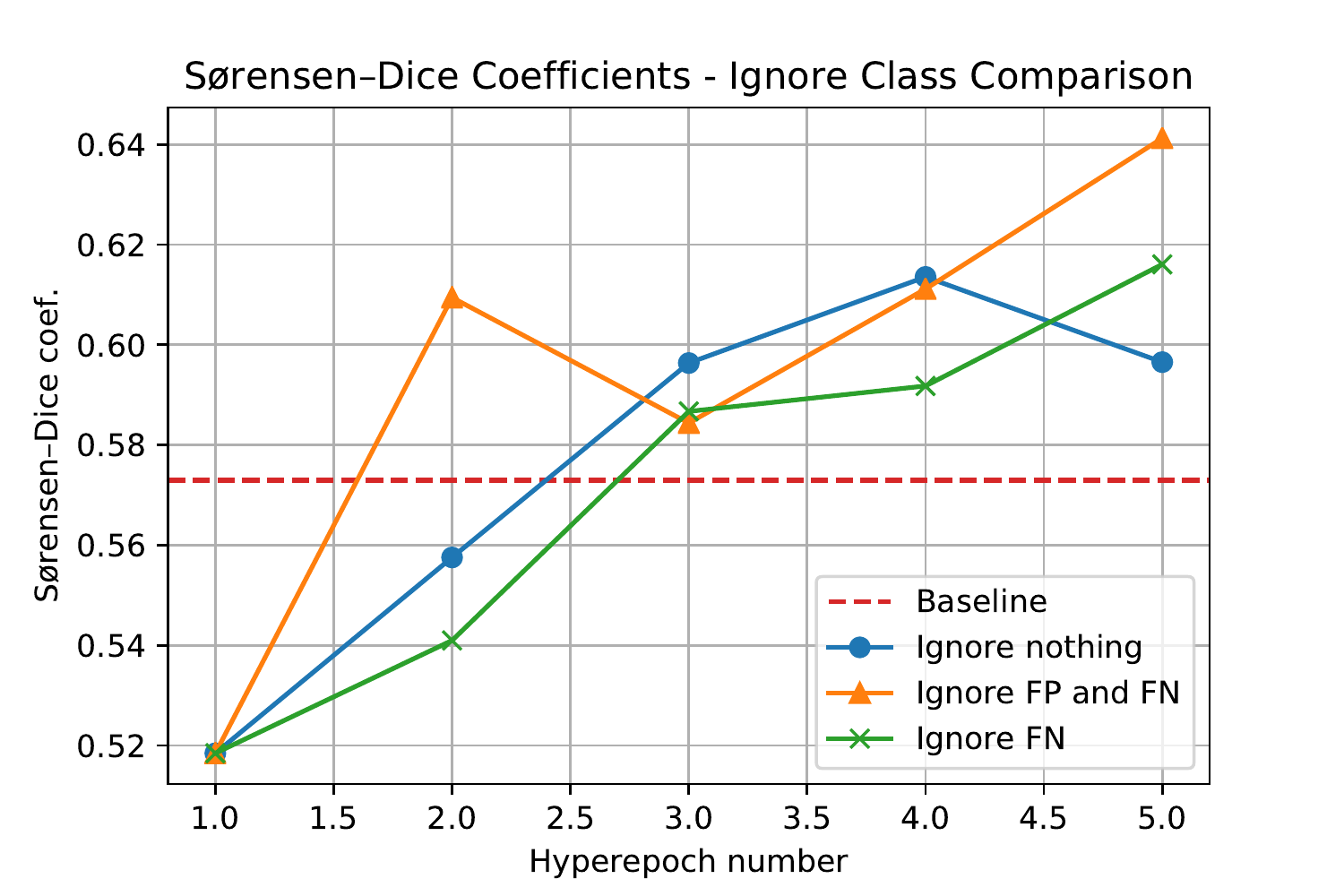}\\
        (a)
    \end{minipage}
    \break
    \begin{minipage}{0.8\linewidth}
        \centering
        \includegraphics[width=\linewidth]{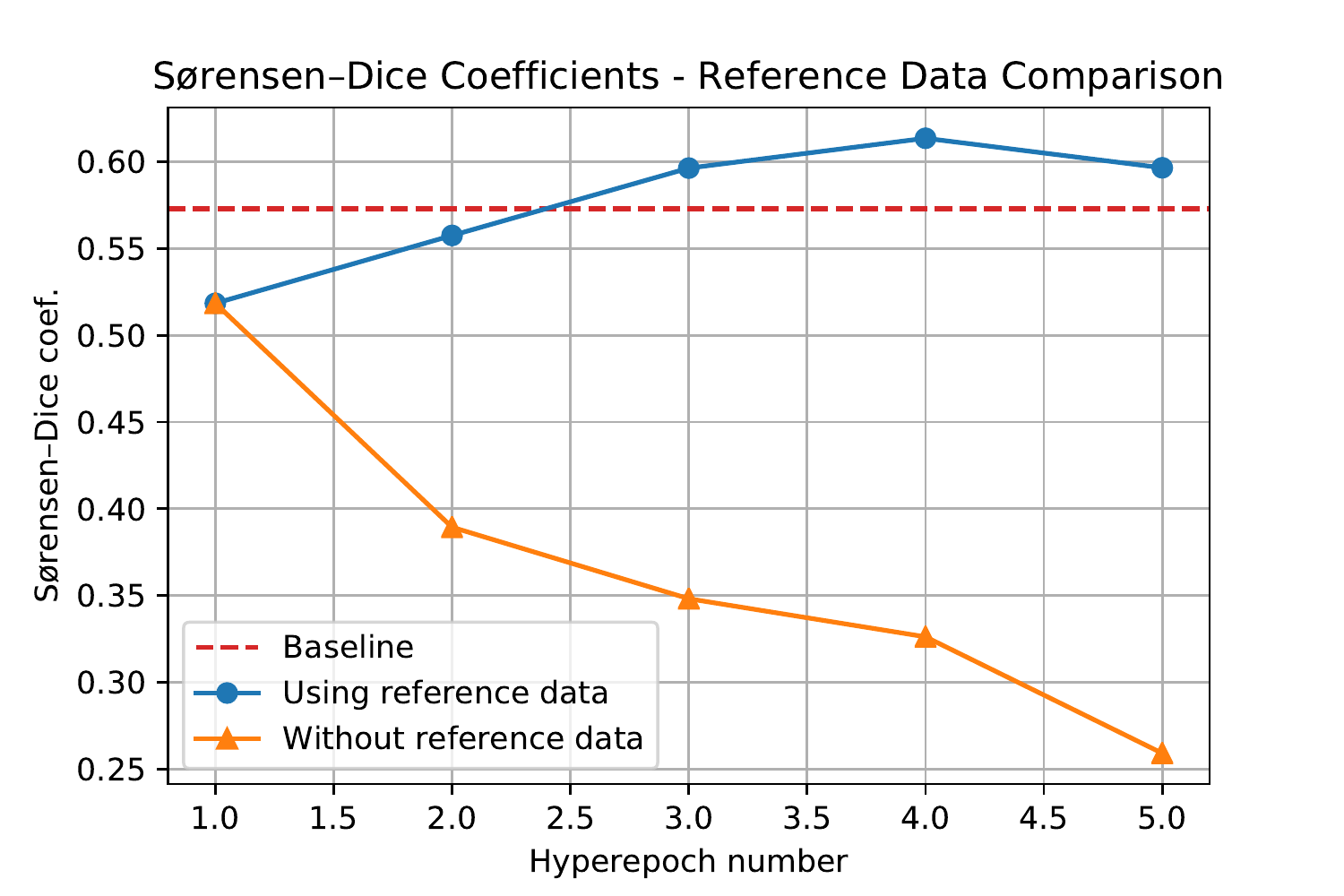}\\
        (b)
    \end{minipage}
    \break
    \begin{minipage}{0.8\linewidth}
        \centering
        \includegraphics[width=\linewidth]{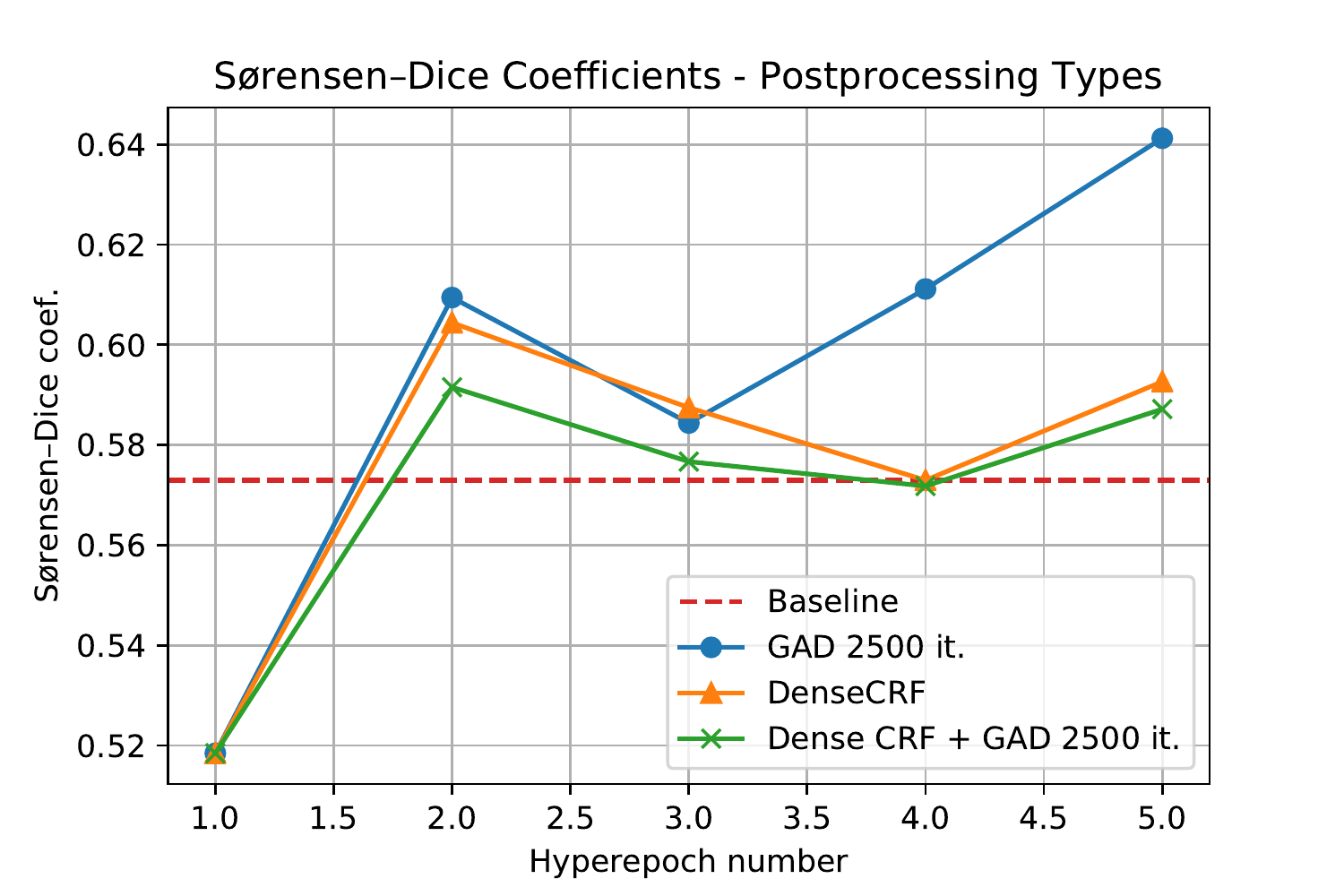}\\
        (c)
    \end{minipage}
    \break%
    
    \caption{Ablation studies. (a) Comparison between strategies for merging network predictions and reference data. (b) Comparison between iterative training with and without the usage of original reference data. (c) Comparison between GAD and Dense CRF.}
    \label{fig:ablation}
\end{figure}

We applied the GAD algorithm to the predictions from a network trained directly on the reference data from HRSCD to evaluate its performance. In Fig.~\ref{fig:gad_122} there is an example of the obtained results. As noted before, we can see in (c) that the change is detected but unchanged pixels around it are also classified as changes by the network. In (d)-(f) it can be clearly seen how the GAD algorithm improves the results by diffusing the labels across similar pixels while preserving edges from the input images in the semantic segmentation results. As expected, more iterations of the algorithm lead to a stronger erosion of incorrect labels. For these results and all others in this section, GAD was applied with $K = 5$ and $\lambda = 0.24$. In Fig.~\ref{fig:gad_vs_dcrf} we can see a comparison between GAD and the Dense CRF\footnote{\url{https://github.com/lucasb-eyer/pydensecrf}} algorithm~\cite{krahenbuhl2011efficient}. While the non-local nature of fully connected CRFs is useful in some cases, we can see the results are less precise and significantly noisier than the ones obtained by using GAD.

To perform quantitative analysis of results, it would be meaningless to use the test data in the HRSCD dataset given that we are attempting to perform a task which is not the one for which ground truth data are available, \ie we are attempting to perform pixel-level precise change detection and not parcel-level change detection. For this reason we have manually annotated the changes as precisely as possible for two 10000x10000 image pairs in the dataset, for a total of 2$\cdot$10$^{8}$ test pixels, or 50~km$^2$. The image pairs were chosen before any tests were made to avoid biasing the results. Due to the class imbalance, total accuracy, \ie the percentage of correctly classified pixels, provides us with a skewed view of the results biased towards the performance on the class more strongly represented. Therefore, the S{\o}rensen-Dice coefficient (equivalent to the F1 score for binary problems) from the point of view of the change class was used~\cite{dice1945measures, sorensen1948method}. The S{\o}rensen-Dice coefficient score is defined as
\begin{equation}
    \mathit{Dice} = (2\cdot TP)/(2\cdot TP + FP + FN)
\end{equation}
where TP means true positive, FP means false positive, and FN means false negative. It serves as a balanced measurement of performance even for unbalanced data.

All tests presented here were done using PyTorch~\cite{paszke2017automatic}. At each hyperepoch, the network was trained for 100 epochs with an ADAM algorithm for stochastic optimization~\cite{kingma2014adam}, with learning rate of $10^{-3}$ for the first 75 epochs and $10^{-4}$ for the other 25 epochs. The tests show the performance of networks trained with the proposed method for 5 hyperepochs (iterations of training and cleaning the data), where the first one is done directly on the available data from the HRSCD dataset. For accurate comparison of methods and to minimize the randomness in the comparisons, the obtained network at the end of hyperepoch 1 is used as a starting point for all the methods. This ensures all networks have the same initialization at the point in the algorithm where they diverge. A baseline network was trained for the same amount of epochs and hyperepochs but with no changes done to the training data. This serves as a reference point as to the performance of the fully convolutional network with no weakly supervised training methods.

\begin{figure*}[t]
    \centering
    \begin{minipage}{0.15\linewidth}
        \centering
        \includegraphics[width=\linewidth]{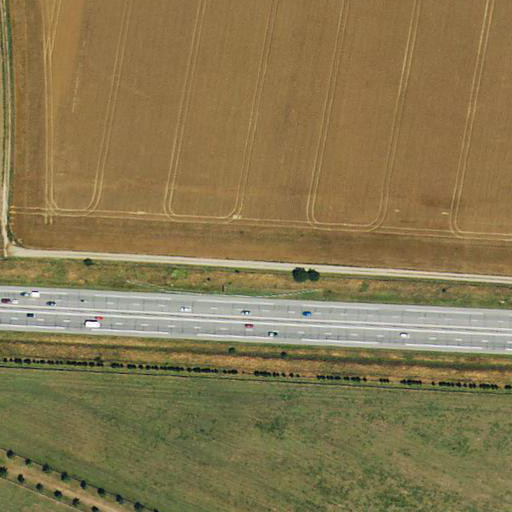}\\
        (a) Image 1
    \end{minipage}
    \hfill
    \begin{minipage}{0.15\linewidth}
        \centering
        \includegraphics[width=\linewidth]{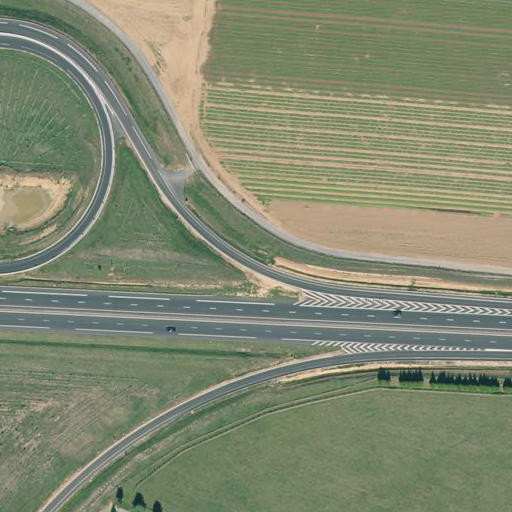}\\
        (b) Image 2
    \end{minipage}
    \hfill
    \begin{minipage}{0.15\linewidth}
        \centering
        \includegraphics[width=\linewidth]{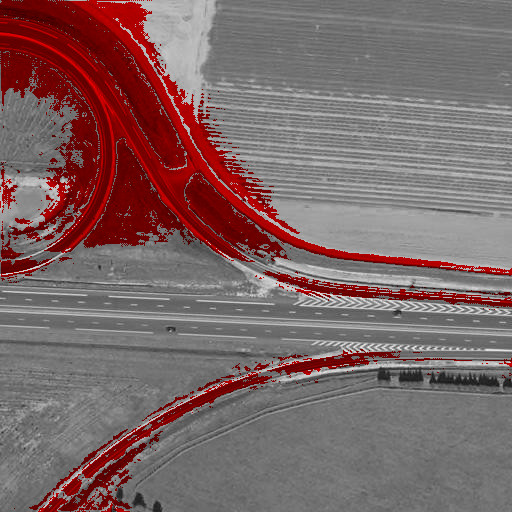}\\
        (c) Baseline
    \end{minipage}
    \hfill
    \begin{minipage}{0.15\linewidth}
        \centering
        \includegraphics[width=\linewidth]{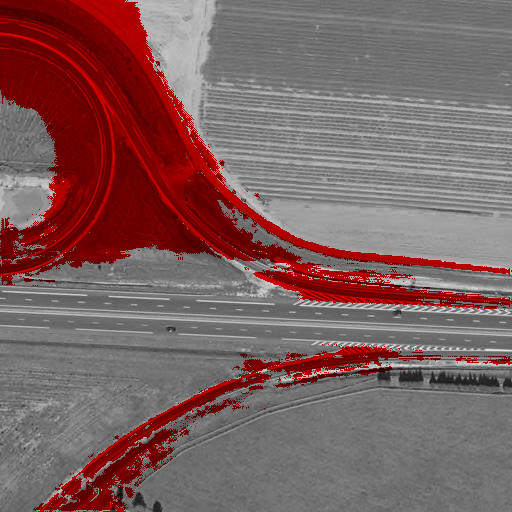}\\
        (d) GAD 2500 it.
    \end{minipage}
    \hfill
    \begin{minipage}{0.15\linewidth}
        \centering
        \includegraphics[width=\linewidth]{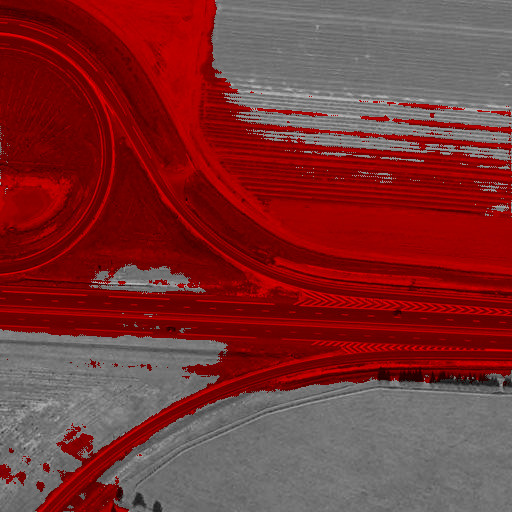}\\
        (e) No ref. const.
    \end{minipage}
    \hfill
    \begin{minipage}{0.15\linewidth}
        \centering
        \includegraphics[width=\linewidth]{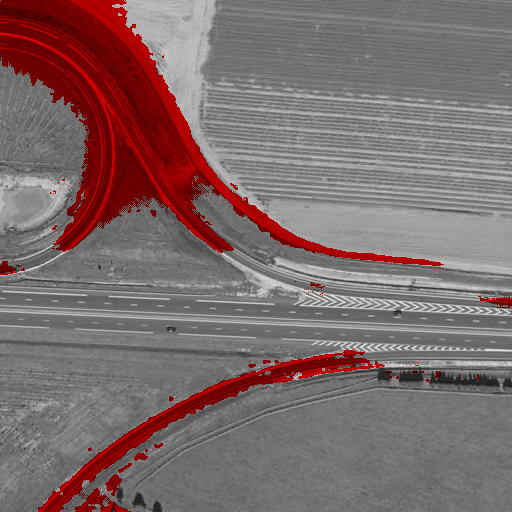}\\
        (f) Dense CRF
    \end{minipage}
    \break%
    
    \begin{minipage}{0.15\linewidth}
        \centering
        \includegraphics[width=\linewidth]{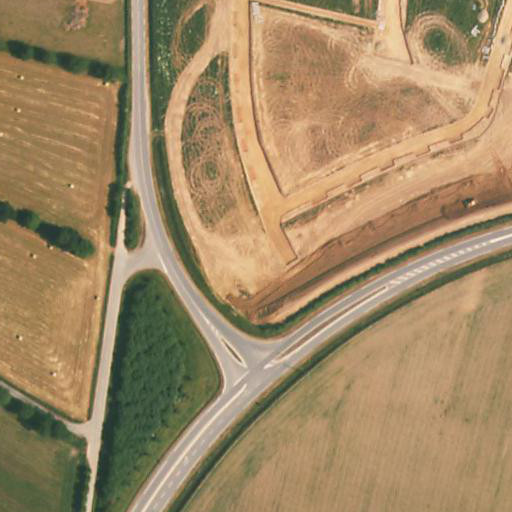}\\
        (g) Image 1
    \end{minipage}
    \hfill
    \begin{minipage}{0.15\linewidth}
        \centering
        \includegraphics[width=\linewidth]{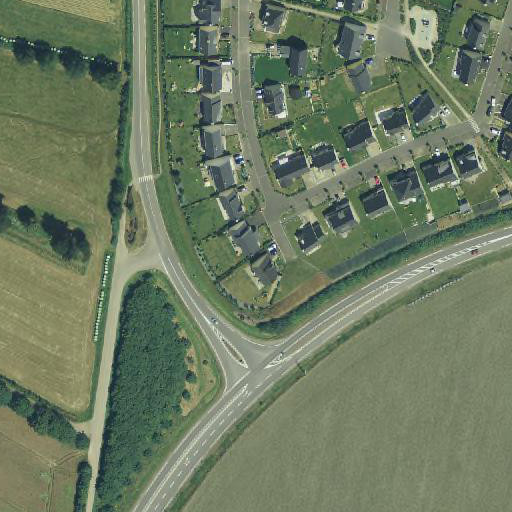}\\
        (h) Image 2
    \end{minipage}
    \hfill
    \begin{minipage}{0.15\linewidth}
        \centering
        \includegraphics[width=\linewidth]{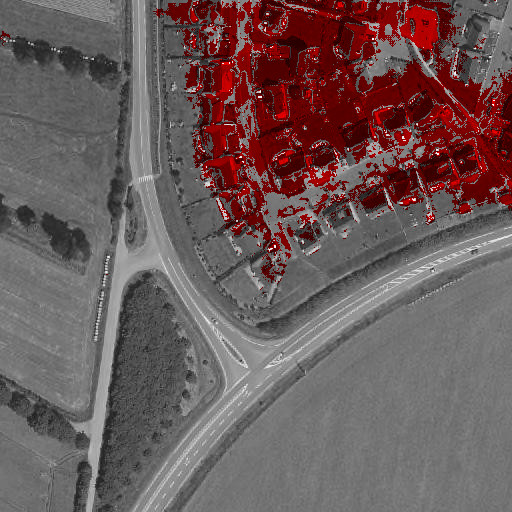}\\
        (i) Baseline
    \end{minipage}
    \hfill
    \begin{minipage}{0.15\linewidth}
        \centering
        \includegraphics[width=\linewidth]{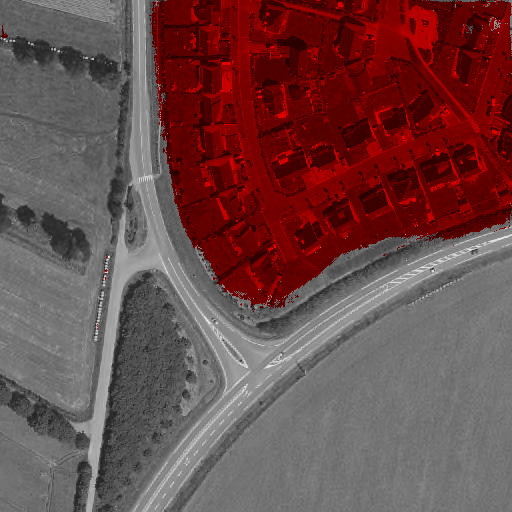}\\
        (j) GAD 2500 it.
    \end{minipage}
    \hfill
    \begin{minipage}{0.15\linewidth}
        \centering
        \includegraphics[width=\linewidth]{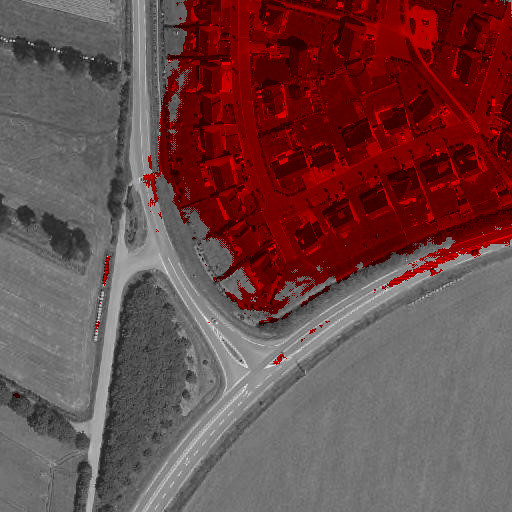}\\
        (k) No ref. const.
    \end{minipage}
    \hfill
    \begin{minipage}{0.15\linewidth}
        \centering
        \includegraphics[width=\linewidth]{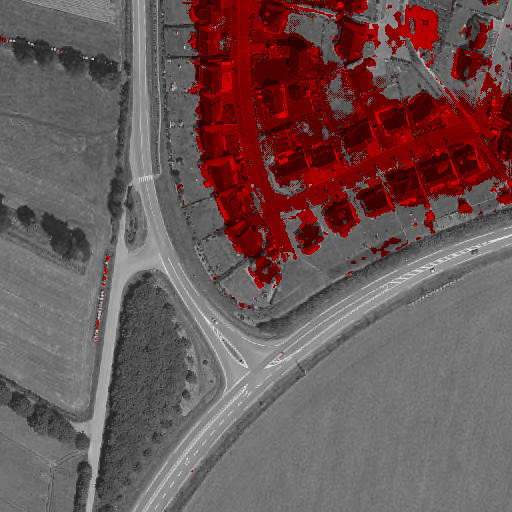}\\
        (l) Dense CRF
    \end{minipage}
    \break%
    
    \caption{Change maps obtained by using different methods on two image pairs. Detected changes are marked in red color.}
    \label{fig:examples}
\end{figure*}

The first comparison, shown in Fig.~\ref{fig:ablation}(a), compares the three methods proposed in Section~\ref{sec:its} to combine the network predictions with the original ground truth from the HRSCD dataset. We notice that all three strategies surpass the baseline network using the proposed iterative training method, which validates the ideas presented earlier. In Fig.~\ref{fig:ablation}(b) we see a comparison between a training using the full training scheme proposed in this paper (without the usage of an ignore class) and the same method but without using the original reference data, \ie using only network predictions processed by GAD to continue training at each hyperepoch. Our results, which corroborate the ones in \cite{khoreva2017simple}, show that referring back to the original data at each hyperepoch is essential to avoid a degradation in performance.

In Fig.~\ref{fig:ablation}(c) we show a comparison between using the proposed GAD algorithm versus the Dense CRF~\cite{krahenbuhl2011efficient} algorithm in the iterated training procedure, as well as using both together. We see that using the Dense CRF algorithm to process predictions leads to good performance in early hyperepochs, but is surpassed by GAD later on. This is likely explained by the non local nature of Dense CRF and its ability to deal with larger errors, but its inferior performance relative to GAD for finer prediction errors.

Figure~\ref{fig:examples} shows the predictions by networks trained by different methods on two example images. We see that the best results are obtained by using the full training scheme with GAD in (d)/(j), followed by Dense CRF, which also achieves good results shown in (f)/(l). The baseline results in (c)/(i), obtained by naively training the network in a supervised manner, and the ones without using the reference data as constraint in the iterative training scheme shown in (e)/(k) are significantly less accurate than those using GAD or Dense CRF.

\section{Analysis}

One possible criticism of the proposed iterative training method is that it would get rid of hard and important examples in the training dataset. It is true that the performance of this weakly supervised training scheme would likely never reach that of one supervised with perfectly clean data, but the results in Section~\ref{sec:experiments} show that using the proposed method we can consistently train networks that perform better than those naively trained with noisy data directly.

The results also made clear that it is of paramount importance to refer back to the ground truth data every time the training ground truth is being modified. Not doing so leads to a fast degradation in performance, since the network simply attempts to learn to copy itself and stops learning useful operations from the data. The results also showed that separating dubiously labelled pixels leads to a small increase in performance, likely due to the fact that we end up providing a cleaner and more trustworthy dataset at training time.

The guided anisotropic diffusion algorithm was compared against the Dense CRF algorithm for using information from the input images to improve semantic segmentation results. While both algorithms were successful when used in the proposed iterative training scheme, GAD outperformed Dense CRF at later hyperepochs for quantitative metrics. Both algorithms yielded visually pleasing results, each performing better in different test cases.

\section{Conclusion}

In this paper we have proposed an iterative training method for training with noisy data that alternates between training a fully convolutional network and leveraging its predictions to clean the training dataset from mislabelled examples.
We showed that the proposed method outperforms naive supervised training using the provided reference data for change detection. We proposed three methods for merging network predictions with reference data, the best of which aimed to ignore suspiciously labelled examples. Our results corroborated previous results which stated that referring back to reference data when performing classification filtering for data cleaning.
We also proposed the guided anisotropic diffusion algorithm for improving semantic segmentation results by performing a cross image edge preserving filtering. The GAD algorithm was used in conjunction with the iterative training method to obtain the best results in our tests. The GAD algorithm was compared against the Dense CRF algorithm, and was found to be superior in performance when used with the proposed iterative training scheme.

\newpage
{\small
\bibliographystyle{ieee}
\bibliography{refs}
}

\end{document}